%% file: emnlp2021.tex
\pdfoutput=1

\documentclass[11pt]{article}

\usepackage[final]{emnlp2021}
\usepackage{graphicx}
\usepackage{times}
\usepackage{latexsym}
\usepackage{pifont}
\usepackage{amssymb}
\usepackage{amsmath}
\usepackage{booktabs}
\usepackage{multirow}

\usepackage{makecell}
\usepackage[table]{xcolor}

\usepackage[T1]{fontenc}

\usepackage[utf8]{inputenc}

\usepackage{microtype}
\usepackage{amsmath}

\title{Plugging Schema Graph into Multi-Table QA: \\ A Human-Guided Framework for Reducing LLM Reliance}

\author{
\begin{tabular}{c}
Xixi Wang$^{1}$ \quad Miguel Costa$^{1}$ \quad Jordanka Kovaceva$^{2}$ \\
Shuai Wang$^{2}$ \quad Francisco C. Pereira$^{1}$
\end{tabular} \\
\\
$^{1}$Technical University of Denmark, Kgs. Lyngby, Denmark \\
$^{2}$Chalmers University of Technology, Gothenburg, Sweden \\
\texttt{s232253@dtu.dk}, \texttt{migcos@dtu.dk}, \texttt{jordanka.kovaceva@chalmers.se}, \\
\texttt{shuaiwa@chalmers.se}, \texttt{camara@dtu.dk}
}

\begin{document}
\maketitle
\begin{abstract}
Large language models (LLMs) have shown promise in table Question Answering (Table QA). However, extending these capabilities to multi-table QA remains challenging due to unreliable schema linking across complex tables. Existing methods based on semantic similarity work well only on simplified hand-crafted datasets and struggle to handle complex, real-world scenarios with numerous and diverse columns. To address this, we propose a graph-based framework that leverages human-curated relational knowledge to explicitly encode schema links and join paths. Given a natural language query, our method searches on graph to construct interpretable reasoning chains, aided by pruning and sub-path merging strategies to enhance efficiency and coherence. Experiments on both standard benchmarks and a realistic, large-scale dataset demonstrate the effectiveness of our approach. To our knowledge, this is the first multi-table QA system applied to truly complex industrial tabular data.
\end{abstract}

\section{Introduction}

Table Question Answering (Table QA) involves answering natural language questions using structured tabular data. This capability is increasingly critical in numerous practical domains, including healthcare~\cite{singhal2025toward}, finance~\cite{srivastava2024evaluating}, and transportation~\cite{zhang2024large}, where data is commonly stored in relational databases \cite{wang-yu-2025-iquest} or spreadsheet-like tables. Effective Table QA systems allow non-expert users to query structured data, thus enhancing accessibility, facilitating data-driven decision-making, and reducing the reliance on manual data manipulation or complex query languages such as SQL.

\begin{figure}
    \centering
    \includegraphics[width=1\linewidth]{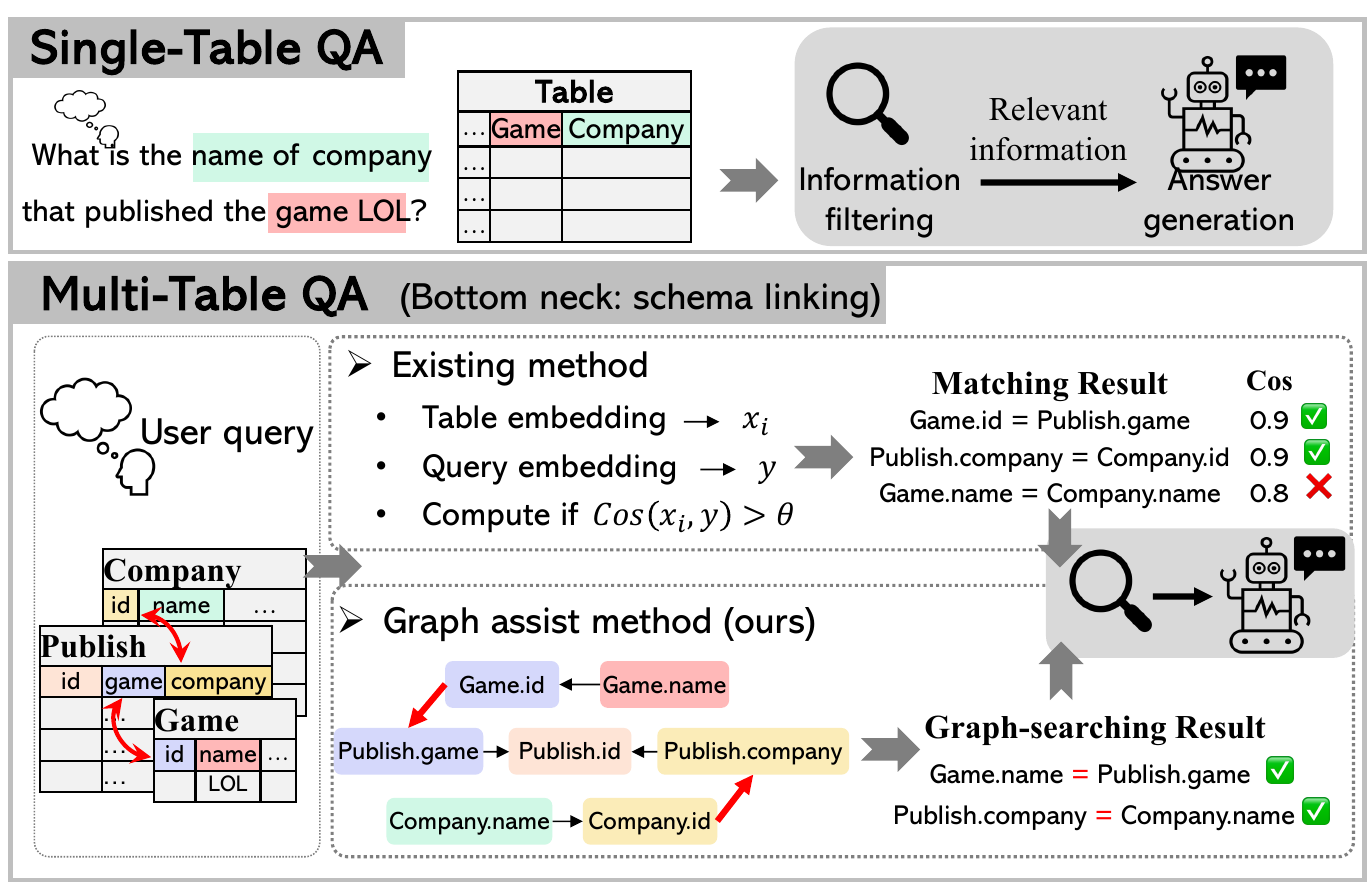}
    \caption{Overview of three Table QA paradigms. \textbf{Top}: Single-table QA, where all relevant information is contained within one table. \textbf{Middle}: Existing multi-table QA approaches that rely on semantic similarity for schema linking across tables. \textbf{Bottom}: Our proposed method, which leverages graph-based, human-curated relational knowledge to explicitly guide schema linking and inter-table reasoning.}
    \label{fig:enter-label-intro}
\end{figure}

Unlike traditional text-based QA~\cite{jin2022survey}, Table QA requires models to understand the structural semantics of tabular data, including column types, row-level content, and inter-cell relationships. In single-table QA, all relevant information resides within one table (see the top of Figure~\ref{fig:enter-label-intro}). Existing methods typically solve this by either translating natural language questions into executable programs (e.g., SQL via semantic parsing)~\cite{nan2022fetaqa, liutapex, cai2022star}, or by treating the table as unstructured input for neural models~\cite{herzig2020tapas, zhu2021tat, ma2022open, pal2022parameter}.

Multi-table QA introduces greater complexity, as answering a question often requires reasoning over multiple related tables (middle of Figure~\ref{fig:enter-label-intro}). This setting poses unique challenges such as identifying relevant tables, aligning schemas, and inferring join paths, which are not present in single-table QA. Existing methods attempt to address these through flattening and concatenating multiple tables~\cite{pal2023multitabqa, wang2024multi}, leveraging key constraints for semantic alignment~\cite{chen-etal-2024-table}, or adopting multi-step retrieval pipelines~\cite{wu-etal-2025-mmqa}. 

However, existing multi-table QA methods have been evaluated only on simplified benchmarks. For instance, the Multi-table and Multi-hop Question Answering (MMQA) benchmark~\cite{wu-etal-2025-mmqa} contains tables with an average of just five columns. These are far less complex than real-world datasets like Crash Investigation Sampling System (CISS)~\cite{nhtsa-2024-ciss}, where tables can have up to 207 columns. These methods typically rely on heavy model training to learn schema linking or semantic alignment, which may suffice for small benchmarks but becomes highly impractical for large-scale tables. Even state-of-the-art LLMs struggle with accurate column-level matching at this scale~\cite{zhang2024tablellama,zhang2024tablellm}. As schema linking across complex tables remains a major bottleneck, current approaches fail to generalize to realistic industrial scenarios.

We argue that multi-table QA does not necessarily require complex model-based or semantic matching techniques to resolve schema linking across tables. In real-world table QA system, the set of tables and their structures are typically fixed. This allows us to leverage domain knowledge to manually construct an explicit graph that captures both schema linking and join path inference. By doing so, we bypass the major bottleneck of implicit schema alignment during inference. Moreover, this approach enhances transparency and controllability of schema linking, making the overall multi-table QA pipeline more robust and easier to adapt to new domains without needing extensive retraining. Importantly, the graph-based schema representation can be implemented as a plug-and-play module, enabling seamless integration with both closed-source and open-source LLMs.

In this paper, we introduce \textbf{SGAM} (Schema Graph-Assisted Multi-table QA), a graph-based framework that integrates human-curated schema knowledge to support multi-table QA, as shown in Figure~\ref{fig:enter-label-intro} (bottom). Specifically, we construct a relational graph capturing the semantic and logical dependencies among table columns. Each graph node encodes essential schema details, such as column name, associated table, and primary key status, while edges represent both intra-table semantic relationships and inter-table joinable or derivation-based links. Given a natural language query, our method searches directed paths within this graph to derive clear and interpretable reasoning chains, significantly simplifying multi-table reasoning tasks. Furthermore, we introduce efficient path pruning and sub-path merging strategies to minimize redundant joins and enhance inference coherence.

Our work offers three main contributions. First, we propose a novel graph-based framework for multi-table QA that leverages human-curated relational knowledge to explicitly represent schema linking and join paths, thereby greatly simplifying reasoning over complex real-world tables. Second, we build and publicly release a new multi-table QA dataset based on the real-world CISS, which is substantially more realistic and challenging than existing benchmarks. Finally, we achieve state-of-the-art performance on both standard multi-table QA benchmarks and our new dataset. To the best of our knowledge, this is the first application of multi-table QA to truly large-scale, real-world tabular data.

\section{Problem Description}

\input{ProblemDescription}

\section{Method }
\input{Graph-based-QA-system}

\section{Experiments}
\input{evaluation}

\section{Related Work}

\paragraph{Single-Table Question Answering.}
Early work treats table QA as semantic parsing: a question is translated into executable SQL or \texttt{pandas} code that is then run on a single table~\cite{nan2022fetaqa,liutapex,cai2022star}.  
Extractive models instead read the whole table as text and select the answer cell~\cite{herzig2020tapas,zhu2021tat}, while generative models directly produce the answer without an explicit program~\cite{ma2022open,pal2022parameter}.  
Recent studies leverage large language models (LLMs): some guide LLMs to emit higher-quality SQL~\cite{zhang2024tablellama,ye2023large,zhang2024tablellm}, whereas others let the LLM read the table and form the answer end-to-end~\cite{wu2025tablebench}.

\paragraph{Multi-Table Question Answering.} Multi-table QA focuses on answering queries that require reasoning over multiple interrelated tables. A core challenge lies in identifying implicit join paths, i.e., determining which columns across tables are semantically or relationally aligned—since such connections are rarely explicit in the query.
Early work by Pal et al.~\cite{pal2023multitabqa} introduced a large-scale multi-table QA benchmark, where all relevant tables are flattened and concatenated with the question, followed by Transformer-based modeling and multi-table pretraining. Chen et al.~\cite{chen-etal-2024-table} proposed a join-aware method that scores column-level joinability based on name semantics, value overlaps, and schema constraints. Wu et al.~\cite{wu-etal-2025-mmqa} utilized LLMs to decompose complex questions into sub-questions, iteratively retrieving and linking tables based on column overlaps. While effective in controlled settings, these methods struggle with real-world scenarios such as CISS~\cite{nhtsa-2024-ciss}, where tables may contain over 200 columns. The resulting complexity severely hinders accurate table linking and cross-table reasoning.

To address these limitations, we propose a graph-assist approach that encodes column-level relations across tables using schema-derived structure. This enables interpretable and scalable reasoning by selectively retrieving relevant columns based on query intent. To the best of our knowledge, this is the first system evaluated on the realistic tables, demonstrating scalability to real-world industrial data.

\section{Conclusion}
In this work, we present a novel graph-based framework for multi-table question answering that integrates explicit, human-curated relational knowledge to assist in schema linking and reasoning. By constructing a structured relational graph that captures both intra- and inter-table dependencies, our method enables interpretable reasoning paths for complex queries, significantly improving robustness and scalability in real-world settings. We further introduce pruning and sub-path merging strategies to reduce redundancy and enhance inference coherence. To support real-world evaluation, we release a challenging multi-table QA dataset based on the CISS corpus. Experiments on both standard benchmarks and our newly constructed dataset demonstrate the effectiveness of the proposed approach. By offloading structural and semantic understanding to a lightweight schema graph, our framework provides a scalable and interpretable solution that advances the practical deployment of multi-table QA in complex real-world industrial settings.


\section{Limitations}

 \textbf{Limitations of Domain Knowledge}. While our method demonstrates strong performance with open-source models, we do not fine-tune them on domain-specific data. As a result, the models may occasionally produce factual errors due to gaps in vehicle safety domain knowledge. Incorporating domain-adaptive fine-tuning could further improve answer accuracy in these expert settings.

\textbf{Dependency on Accurate Attribute Retrieval}. While our method significantly improves multi-table QA accuracy by using schema-guided reasoning over a graph after attributes retrieval, it still relies on correctly identifying most of relevant attributes during the retrieval phase. If Attribute-RAG fails to extract the very necessary columns corresponding to semantic components of the question, the downstream graph reasoning process may lack critical information, ultimately leading to incorrect or incomplete answers. In future work, we plan to explore more robust and semantically grounded methods for mapping each part of a natural language question to its corresponding variables during the initial retrieval stage.

\section*{Funding sources}
Part of this research was supported by the Chalmers Transport Area of Advance project TREND.
\section*{Acknowledgements}
The authors would like to thank e-Commons (research infrastructure at Chalmers University of Technology) for their consultations, and extend special thanks to Nora Speicher for valuable discussions related to running the experiments. We acknowledge the National Academic Infrastructure for Supercomputing in Sweden (NAISS), partially funded by the Swedish Research Council through grant agreement no. 2022-06725, for awarding this project access to the Alvis  cluster for computations and data handling.

\bibliography{anthology,custom}
\bibliographystyle{acl_natbib}

\appendix

\input{Appendix}

\end{document}

%% file: ProblemDescription.tex
Multi-table question answering involves answering a natural language question by retrieving and reasoning over multiple structured tables.
Let the table corpus be defined as $\mathcal{C} = \{\mathcal{T}_1, \ldots, \mathcal{T}_M\}$, where each table $\mathcal{T}_i$ is represented as $\mathcal{T}_i = \{C_1: A_1, \ldots, C_d: A_d\}$. Here, $C_j$ denotes the name of the $j$-th column and $A_j$ its associated values, with $d$ being the number of columns in $\mathcal{T}_i$.
Given a natural language question $Q$, the system must identify a subset of relevant tables $E(Q) = \{\mathcal{T}_{Q,1}, \ldots, \mathcal{T}_{Q,k}\} \subseteq \mathcal{C}$ and determine how to join them through appropriate conditions $\bowtie_{c_j}$, such as foreign key relationships or semantically inferred links. These joins form a join path:
\begin{equation}
    \mathcal{T}_{Q,1} \bowtie_{c_1} \mathcal{T}_{Q,2} \bowtie_{c_2} \cdots \bowtie_{c_{k-1}} \mathcal{T}_{Q,k}.
\end{equation}

The system must also retrieve a set of relevant attributes $A(Q) = \{a_1, a_2, \ldots, a_n\}$, where each $a_j$ belongs to one of the selected tables, i.e., $a_j \in \mathcal{T}_{Q,j}$. The retrieved attributes and joined tables together must provide sufficient and relevant information to generate a natural language answer $\mathcal{A}$ to query $Q$.

%% file: Graph-based-QA-system.tex
\begin{figure*}
    \centering
    \includegraphics[width=1\linewidth]{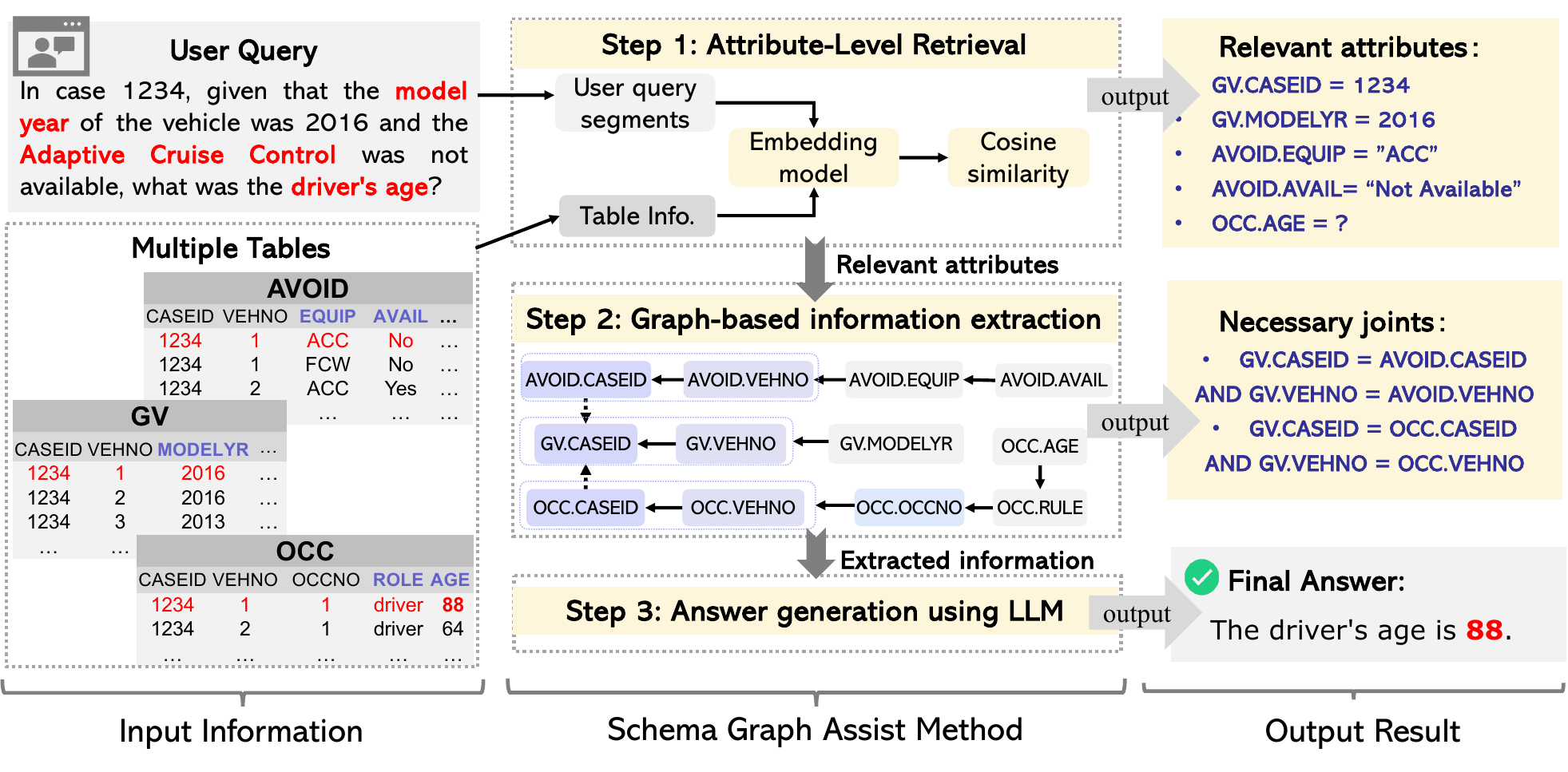}
    \caption{Illustration of our Schema Graph Assist Multi-table QA framework. Given a complex user query and multiple tables, Step 1 retrieves relevant attributes from the tables using embedding-based semantic similarity. Step 2 uses a schema graph to automatically identify and traverse the minimal set of necessary join paths, such as GV.CASEID = AVOID.CASEID and GV.VEHNO = AVOID.VEHNO, to connect related information across tables. Step 3 uses LLM to generate a final answer based on extracted information. This graph-guided reasoning enables the system to accurately extract the target attribute (OCC.AGE) by aligning semantically related fields and resolving multi-hop dependencies. In this case, the system correctly infers the driver’s age as 88.}
    \label{fig:enter-label}
\end{figure*}

\subsection{Attribute-Level Retrieval}

\textbf{Query decomposition:} 
Traditional retrieval-augmented generation (RAG) methods handle natural language queries well but struggle in multi-table settings. Queries often embed multiple implicit constraints (e.g., \textit{vehicle models}, \textit{case IDs}) within a single sentence, making it difficult to identify all relevant filters. This can lead to incomplete or imprecise retrieval. Moreover, connective phrases (e.g., \textit{at the time of}, \textit{related to}) enhance fluency but add little semantic value, introducing noise into the embedding space and increasing the likelihood of irrelevant matches.

To overcome these challenges, we propose a decomposition-based approach, explicitly segmenting each query into two distinct components: \emph{constraints} and \emph{question focus}. The \emph{constraints} specify conditions to filter relevant records (e.g., particular vehicle models or case identifiers), while the \emph{question focus} identifies the core information sought (e.g., activation of safety features, driver's age). Handling these components separately allows more accurate retrieval and reduces interference from non-essential linguistic elements.

\noindent\textbf{Retrieval tabular attributes:}
To support accurate attribute identification, our RAG framework constructs structured keyword sets for each attribute. These keywords, derived from attribute names, descriptions, and representative values, fall into two categories: \textit{uniqueness-based keywords} that uniquely identify an attribute, and \textit{frequency-based keywords} that frequently occur within an attribute’s values but rarely elsewhere. For instance, the term \texttt{animal} is typically associated with variables related to \texttt{CRASHTYPE} (crash type), while terms like \texttt{truck} and \texttt{bus} frequently occur in the \texttt{BODYCAT} (vehicle's body category) attribute. This design enables the retrieval process to harness both semantically distinctive signals and statistically salient patterns.  

Given a query segment \( q \), we embed it alongside each attribute’s keyword set \( K_i \) using a pre-trained embedding model, such as OpenAI's \texttt{text-embedding-3-small}. The cosine similarity between the query and each keyword set is then computed as \( s_i = \cos(\text{emb}(q), \text{emb}(K_i)) \). To better distinguish relevant from irrelevant attributes, similarity scores are squared and normalized:
\begin{equation}
    s_i' = \frac{s_i^2-\min_k s_k^2}{\max_j s_j^2-\min_k s_k^2}
\end{equation}
We select keywords exceeding a threshold  (i.e., \( s_i' \ge \tau \)) and map them back to candidate attributes, forming the retrieval context for subsequent RAG processing. This method effectively balances recall and precision by exploiting both semantic relevance and frequency-based indicators, enhancing retrieval performance on noisy datasets.

\subsection{Graph based information extraction}
\paragraph{Graph Construction.}
We integrate relational database knowledge into our graph construction to preserve semantic structures and relational contexts across multiple tables. Attributes from these tables are represented as nodes within a directed graph $G = (V, E)$. Each node $v \in V$ is uniquely identified by a tuple $v = (C, t, \pi)$, where $C$ is the column name, $t$ the table name, and $\pi$ a boolean indicating whether $C$ is the primary key. For instance, the node ("ACTIVATE", "AVOID", 0) denotes the column \texttt{ACTIVATE} in the \texttt{AVOID} table, not being a primary key.

Edges $e = (v_i, v_j) \in E$ represent attribute dependencies and are classified into intra-table edges and inter-table edges based on their connections.

\paragraph{Intra-table Edges.}
We define intra-table edges to capture both structural keys and semantic relationships among attributes within the same table. These edges fall into two categories:

(1)\noindent\textit{ Primary key-based edges}: Directed edges are created from non-key attributes to primary key attributes within a table, reflecting inherent structural dependencies. For example, the edge from \texttt{("OCCNO", "OCC", 1)} to \texttt{("VEHNO", "OCC", 1)} encodes the association between occupants and their vehicles:
\begin{equation}
E_{\text{intra}}^{\text{key}} = \{(v_i, v_j) \in E \mid v_i.t = v_j.t \land v_j.\pi = 1\}
\end{equation}
(2)\textit{ Semantic context edges}: Certain attributes lack standalone meaning and must be interpreted in combination with others. For instance, \texttt{ACTIVATE} (indicating system activation) is meaningful only when contextualized with \texttt{EQUIP} (the system type) and \texttt{AVAIL} (system availability). To maintain semantic integrity, we manually introduce edges among such interdependent attributes:
\begin{equation}
E_{\text{intra}}^{\text{context}} = \{(v_i, v_j) \in E \mid v_i.t = v_j.t\}
\end{equation}

\noindent\textbf{Inter-table Edges.}
Inter-table edges capture relationships between attributes across different tables, supporting schema linking and multi-hop reasoning. These edges fall into two categories:

(1)\noindent\textit{ Key-based edges}: These connect attributes with identical names (e.g., \texttt{CASEID}, \texttt{VEHNO}) that serve as primary keys in their respective tables, indicating joinable relationships:
\begin{equation}
    \begin{aligned}
E_{\text{inter}}^\text{key} = \{(v_i, v_j) \in E \mid\ & v_i.t \ne v_j.t \land\ v_i.a = v_j.a\\
&  \land v_i.\pi = v_j.\pi = 1 \}
\end{aligned}
\end{equation}
(2)\textit{ Derivation edges}: For attributes whose values are algorithmically derived from attributes from other tables, we explicitly add edges to reflect their relationship. For example, the attribute \texttt{ALCOHOL INVOLVEMENT} in table \texttt{CRASH} is computed based on the values of \texttt{ALCOHOL TEST RESULT FOR DRIVER} in table \texttt{GV}. Therefore, in the graph, the node \texttt{("ALCINV", "GV", 0)} will have outcoming edges to \texttt{("ALCTEST", "GV", 0)}, representing its derivation logic.
\begin{equation}
        E_{\text{inter}}^{\text{derive}} = \{(v_i, v_j) \in E \mid v_i.t \ne v_j.t \land v_i.a = v_j.a \}
\end{equation}

To unify structural and semantic reasoning, we merge intra-table edges and inter-table edges into a single graph structure:
\begin{equation}
    E = (E_{\text{intra}}^{\text{key}} \cup E_{\text{intra}}^{\text{context}})\land (E_{\text{inter}}^{\text{key}} \cup E_{\text{inter}}^{\text{derive}})
\end{equation}
We also manually assign labels for different edges to clearly represent both extraction cues and reasoning logic, forming the backbone for downstream multi-table query interpretation.

\paragraph{Path Search for Attribute Dependency Chains.} 
Given a query involving one or more target attributes, we perform a directed path search over the graph $G$ to identify semantic dependency chains. Each search begins from a query-referenced attribute node and proceeds along dependency edges until reaching a terminal node, typically a primary key. For tables with multiple primary keys, we define a single terminal key $v^*$ as the most global (i.e., top-level) anchor point:
\begin{equation}
    v^* = \arg\max_{v_k \in \Pi} \; \text{depth}(v_k)
\end{equation}
where \( \Pi \) is the set of a table's primary keys and \( \text{depth}(\cdot) \) measures the topological distance from the key node to the attribute node. 
This guarantees that the selected path terminates at the most globally joinable key. Each resulting path $P = (v_1, v_2, \ldots, v_n)$ represents a coherent attribute dependency chain. The full answer is computed by aggregating all valid paths:
\begin{equation}
    Answer(Q) = \phi(\mathbb{P}), \ \mathbb{P} = \{P_1, P_2, \ldots, P_n\}
\end{equation}

\paragraph{Path Pruning and Sub-path Merging.} 
To reduce redundancy and improve reasoning efficiency, we introduce a pruning and merging strategy over identified attribute dependency paths. For each target attribute node $v_i \in T_j$, the graph search typically yields multiple dependency paths $P(v_i, v_c)$ ending at a terminal key node $v_c$ (often a primary key) in either the same or different tables. These paths may span within or across tables, leading to overlapping semantics and redundant retrieval when they converge on the same key.


To address this, we apply the following pruning procedure to each path in $\mathbb{P}$. For a path $P$, let $e^*$ denote the last inter-table edge from the attribute node. We truncate the path to end at $e^*$, retaining only the sub-path $P' \subseteq P$ that lies entirely within a single table:
\begin{equation}
    P' = P(e^*.\text{dst}, v_c), \ e^*.\text{dst}.t = v_c.t
\end{equation}
where $e^*.dst$ denotes the target node of the last inter-table edge. This ensures each retained sub-path  $P'$ stays within a single table and links a relevant attribute to its terminal key. Merging these $P'$ yields a minimal, non-redundant set of attribute-to-key chains, avoiding duplicate extraction, maintaining contextual coherence, and reducing computational cost

%% file: evaluation.tex
We evaluate our graph-based method through two experiments: SQL generation on the BIRD benchmark and end-to-end QA on the real-world CISS dataset. We now detail the setup and results~\footnote{https://github.com/hiXixi66/SGAM-Multi-table-QA}.

\subsection{Datasets}

\noindent\textbf{BIRD:}
To evaluate the effectiveness of approach in multi-table information retrieval, we first used the public BIRD\footnote{https://bird-bench.github.io/} dataset \citep{li2024can}. BIRD is composed of 95 table corpus ranging from healthcare, transportation, sports, and retail and each corpus contains 7.4 tables on average. From BIRD, we selected two complex corpus available: \textit{Olympics} and \textit{Financial}, because BIRD was originally designed for text-to-SQL tasks, rather than multi-table question answering. The difference in task settings means that the full dataset is not directly compatible with our method, which requires graph construction for multi-table reasoning. Therefore, we manually constructed schema graphs for two representative and complex subsets, \textit{Olympics} and \textit{Financial}, to adapt them to our QA framework. \textit{Olympics} contains 11 tables with an average of 2.9 attributes per table and Financial is formed by 8 tables with and average of 6.7 attributes per table. For each corpus, we built individual graphs corresponding to the specific domain.

\noindent\textbf{CISS:}
To showcase the real-world applicability of our approach, we construct a new multi-table question answering dataset based on the CISS. All tables in this dataset are domain-specific and task-relevant, with well-defined key relationships. 
CISS provides in-depth crash data from the United States, including detailed information on crash types, accidents' causes, injury severity, vehicle types, and more. The dataset covers approximately 3000 to 4500 crashes annually. The recorded crash cases span from accident year 2017 to 2023.
We began by converting structured tabular data into textual summaries and use GPT-4o to automatically generate question-answer pairs. Each question is manually verified for clarity, relevance, and correctness, resulting in a verified accuracy of 98\%. The final dataset consists of more than 620 high-quality QA pairs covering diverse aspects of traffic accidents, including crash types, vehicle conditions, driver and pedestrian demographics, and injury outcomes. To encourage varied reasoning complexity, the dataset includes 220 one-hop, 300 two-hop, and 100 multi-hop questions, where a \textit{hop} refers to a distinct reasoning step required to connect different pieces of information across attributes or tables. This CISS-based QA dataset also serves as a case study to evaluate the effectiveness of our graph-enhanced approach in complex, domain-driven applications.



%
\begin{table}[t]
\footnotesize
\centering
\resizebox{\linewidth}{!}{%
\begin{tabular}{ccccc}
\toprule
\multirow{2}{*}{\textbf{Properties}} & \multicolumn{3}{c}{\textbf{Benchmark}} & \textbf{Real Crash} \\
\cmidrule(lr){2-4} \cmidrule(lr){5-5}
 & \textbf{Olympics} & \textbf{Financial}  & \textbf{BIRD} & \textbf{CISS} \\
\midrule
Total Tables               & 11    & 8   & 7.4  & 38\\ 
\midrule
\rowcolor{gray!10} \multicolumn{5}{l}{\textit{Properties per table}} \\
\midrule
Avg attributes/columns     & 2.9    & 6.7 & -  & \textbf{30.2} \\
Avg foreign keys           & 1.8    & 2.3 & -  & 2.1 \\
Avg primary keys           & 1.0    & 1.0 & -  & 2.1 \\
\midrule
\rowcolor{gray!10} \multicolumn{5}{l}{\textit{Question Category}} \\
\midrule
Numerical                  & 78     & 46  & -  & 5 \\
List                       & 71     & 39  & -  & 75 \\
Select                     & 11     & 12  & -  & 548 \\
Count                      & 9      & 9   & -  & 0 \\
\midrule
\rowcolor{gray!10} Total   & 169    & 101 & 12751 & 628 \\
\bottomrule
\end{tabular}
}
\caption{Descriptive statistics and question types for benchmark datasets (Olympics, Financial, BIRD) and the real-world CISS dataset. Notably, CISS tables contain significantly more attributes per table compared to the benchmark datasets.}
\vspace{-0.3cm}
\label{tab:olympics-weather-stat}
\end{table}

\subsection{Baselines}
To ensure an up-to-date comparison on the BIRD dataset, we selected the highest ranking models from the BIRD leaderboard which tracks submitted results proposed in recent literature. All the selected baseline models use GPT-4o as backbone and demonstrate diverse strategies for SQL generation and multi-table reasoning. We compare our results against Distillery+GPT-4o \citep{maamari2024death}, RSL-SQL+GPT-4o \citep{cao2024rsl}, OpenSearch-SQL \citep{xie2025opensearch}, and AskData+GPT-4o (AT\&T CDO-DSAIR), with the latter ranking first among all. The first three methods are all among the top-performing systems on the leaderboard and rely on oracle knowledge, which significantly simplifies the task. AskData+GPT-4o also reports results without oracle knowledge, providing a more realistic comparison scenario.

\subsection{Implementation details}
For embedding-based retrieval, we use the \texttt{text-embedding-3-small} model made available by OpenAI \citep{openai2024embedding}. For model inference, we use LLaMA-3 70B \citep{meta2024llama3}, GPT-3.5 \citep{openai2023gpt35}, GPT-4o \citep{openai2024gpt4o}, Qwen2.5-7B \citep{qwen2.5}, and DeepSeek-R1-Distill-Qwen-32B \citep{deepseekai2025deepseekr1incentivizingreasoningcapability} models on a high-performance computing cluster with 4 NVIDIA A100 GPUs (40GB). All main prompts used throughout the experiments are listed in the Appendix~\ref{sec:appendix_prompt}.


\subsection{Tasks and Evaluation Metrics}
We conduct two types of evaluation using each of the aforementioned datasets.

\noindent\textbf{Graph-aware evaluation (using BIRD):} Designed to directly assess the impact of introducing our graph-based schema representation into SQL generation process, we compare model performance with and without graph guidance to understand how structured schema information influences query accuracy. We report execution accuracy and SQL generation metrics, precision, recall and F1 across different datasets for both variants.

\noindent\textbf{End-to-end evaluation (using CISS):}
Unlike traditional Text2SQL tasks that assume clean, well-structured data, the CISS dataset presents real-world challenges such as mixed structured and textual fields, inconsistent values, and complicated connected tables. Many attributes can only be used when combined with other attributes, and reasoning often goes beyond simple table joins. To demonstrate that a schema graph can be used beyond guide SQL generation tasks, such as resolving inconsistencies and modeling cross-table semantics, we conduct experiments on a real-world QA task over CISS.

We assess each generated answer by evaluating whether it correctly addresses the question’s core information need, and report performance across different hop categories (1-hop, 2-hop, and $\geq$3-hop) to assess reasoning complexity. The backbone models vary across experiments, but the question decomposer model (Q-d) remains fixed per block to isolate the impact of the encoder. No oracle evidence is provided; models must retrieve and reason over raw table data guided by graph-enhanced or baseline retrieval mechanisms.
We use GPT-4o to automatically judge whether the generated answer correctly addresses the question, based on semantic correctness. To ensure reliability, we manually reviewed 30\% of the results, confirming strong agreement between the model’s judgment and human evaluation.

\subsection{Results: Graph-Aware Evaluation}

To evaluate the effectiveness of our proposed graph-based approach for assisting large language models in SQL generation, we conduct experiments on the BIRD dataset. Our task is framed as multi-table question answering, where the model must interpret a question and retrieve relevant information across complex relational schemas. Unlike prior work that assumes access to oracle knowledge---i.e., explicit evidence annotations such as \texttt{Charter schools }refers to \texttt{Charter School (Y/N)` = 1} in table \texttt{fprm}---our setting reflects real-world applications, where such privileged mappings are typically not readily available. To simulate this scenario, we do not provide any evidence annotations to the model in our approach. Instead, we focus on complex corpus that involve multiple interrelated tables and construct a relational graph for each topic to encode schema and semantic dependencies. Questions are grouped by topic and then split into three subsets: 20 examples for training, 30 examples for development, and the remaining for testing.

As shown in the upper part of Table~\ref{tab:model-metrics-transposed}, our method achieves strong execution accuracy (EX), reaching 83.19\% at pass@1 and up to 87.61\% at pass@10 for the \textit{Olympics} dataset and up to 78.57\% for the \textit{Financial} dataset. The average is higher than AskData + GPT-4o, which didn't use oracle knowledge as ours. Importantly, most of baseline methods listed in the lower part of the table depend on oracle annotations and yield lower performance. These results highlight the effectiveness of our human knowledge graph-guided framework in enabling LLMs to generate accurate SQL in a fully automated and without explicit oracle knowledge.
\begin{table}[t]
\centering
\small
\resizebox{\linewidth}{!}{%
\begin{tabular}{p{1.3cm}p{0.8cm}|c|cc}
\toprule
\setlength{\tabcolsep}{2pt}
\multirow{2}{*}{\textbf{Model}} & \multirow{2}{*}{} & \multirow{2}{*}{\makecell[c]{\textbf{Oracle}\\\textbf{Knowledge}}} & \multicolumn{2}{c}{\textbf{Execution on}} \\
 & & &\multicolumn{2}{c}{\textbf{Eval Datasets}} \\
\midrule 
\rowcolor{gray!15} \multicolumn{3}{l}{\textbf{Comparative Method} (pass@1)} & \textbf{Dev } & \textbf{Test }\\
\midrule

\multicolumn{2}{l|}{\makecell[l]{Distillery + GPT-4o\\\citet{maamari2024death} }} & \textcolor{green!60!black}{\ding{51}} & {67.21$^{*}$}  & {71.83$^{*}$} \\

\multicolumn{2}{l|}{\makecell[l]{RSL-SQL + GPT-4o\\\citet{cao2024rsl} }}& \textcolor{green!60!black}{\ding{51}} &{68.12$^{*}$}  & {70.21$^{*}$}   \\

\multicolumn{2}{l|}{\makecell[l]{OpenSearch-SQL \\\citet{xie2025opensearch}}}  & \textcolor{green!60!black}{\ding{51}} & {69.30$^{*}$}  & {72.28$^{*}$} \\
\midrule
 \multicolumn{2}{l|}{\makecell[l]{AskData + GPT-4o  \\(AT\&T CDO-DSAIR) }\footnotetext{This data is collected from: https://bird-bench.github.io/}} & \textcolor{green!60!black}{\ding{51}} & {75.36$^{*}$}& {77.14$^{*}$} \\
 \multicolumn{2}{l|}{\makecell[l]{AskData + GPT-4o  \\(AT\&T CDO-DSAIR) }\footnotetext{This data is collected from: https://bird-bench.github.io/}} & \textcolor{red}{\ding{55}} & {65.91$^{*}$ (\textcolor{red}{\textbf{-9.45}})}& {67.41$^{*}$ (\textcolor{red}{\textbf{-9.73}})} \\

\midrule
\rowcolor{gray!15} \multicolumn{3}{l}{\textbf{Our Method}} & \textbf{Olympics } & \textbf{Financial }\\
SGAM+&pass@1 & \textcolor{red}{\ding{55}} &\cellcolor{green!20}83.19  &  \cellcolor{red!26} 55.36 \\
GPT-4o &pass@5 & \textcolor{red}{\ding{55}} & \cellcolor{green!30}86.73 &  \cellcolor{green!12} 73.21 \\
& pass@10 & \textcolor{red}{\ding{55}} & \cellcolor{green!40}87.61 &  \cellcolor{green!26} 78.57  \\

\bottomrule
\end{tabular}
}
\caption{Execution accuracy on the BIRD Text2SQL benchmark. Scores with * are from the official leaderboard. Since BIRD covers a wide range of domains, we select two representative domains for evaluation. Method \textit{AskData.} shows a large performance drop without Oracle Knowledge (schema linking hints), while our method performs well without using it. Green highlights indicate our method outperforms non-Oracle baselines.}
\label{tab:model-metrics-transposed}
\end{table}


\begin{table*}[t]
\centering
\footnotesize
\setlength{\tabcolsep}{10pt}
\renewcommand{\arraystretch}{1.2}
\begin{tabular}{lcccccc}
\toprule
& \multicolumn{2}{c}{\textbf{P} 
}&\multicolumn{2}{c}{\textbf{R}}& \multicolumn{2}{c}{\textbf{F1 }}\\
\cmidrule(lr){2-3} \cmidrule(lr){4-5}\cmidrule(lr){6-7}
           \textbf{Dataset}       & w/    & w/o $\Delta$     & w     & w/o $\Delta$   & w   &w/o $\Delta$ \\
\midrule
Olympics         & ${84.15\pm1.20}$ & \cellcolor{red!26}-50.98 & ${85.71\pm1.47}$ & \cellcolor{red!25}-50.96& ${84.55\pm1.26}$    & \cellcolor{red!26}-51.26 \\
Financial        & ${58.93\pm3.32}$ & \cellcolor{red!20}-40.75 & ${56.43\pm3.36}$ & \cellcolor{red!19}-38.01& ${56.87\pm3.35}$    & \cellcolor{red!20}-39.19 \\
\midrule
Avg              &    71.54   &    -45.87   &   71.07     &   -44.49     &     70.71   &      -45.23  \\
\bottomrule
\end{tabular}
\caption{Performance comparison for precision (P), recall (R), and F1 score (F1). We compare the performance of GPT-4o with and without graph assistance for the selected BIRD datasets.}
\label{tab:graph-ablation2}
\end{table*}

While GPT-4o plays an important role, its performance degrades significantly without the support of the graph structure. To verify it, we compare the models performance with and without graph-assisted guidance in generating SQL. Table \ref{tab:graph-ablation2} reports the precision, recall, and F1 score on the selected BIRD datasets. As shown, using our approach significantly improves performance. On the \textit{Olympics} dataset, F1 score increases by 51.26\%  (from 33.29\% to 84.55\%), and on the \textit{Financial} dataset by 39.19\% points (from 17.68\% to 56.87\%). These significant improvements demonstrate that by using a graph-based approach we can help guide the model to key attribute relationships between different tables. 
\subsection{Results: End-to-end Evaluation}
To validate the applicability of our method in real-world, noisy, and semantically entangled data settings, we further evaluate it on the CISS dataset. 

\begin{table*}[ht]
\centering
\setlength{\tabcolsep}{10pt}
\renewcommand{\arraystretch}{1.2}
\resizebox{1\linewidth}{!}{%
\begin{tabular}{l|ccc|cc|cc|cc}
\toprule
\multicolumn{1}{r|}{\cellcolor{gray!10}{\textbf{Q-d model} }}&\multicolumn{3}{c|}{{\cellcolor{gray!10}{\textit{-} }}}& \multicolumn{2}{c|}{{\cellcolor{gray!10}{\textit{ GPT-4o} }}} &  \multicolumn{2}{c|}{\cellcolor{gray!10}{\textit{LLaMA-3 70B} }}&  \multicolumn{2}{c}{\cellcolor{gray!10}{\textit{ Qwen2.5-32B} }}\\

\cline{2-2}\cline{3-4}\cline{5-6}\cline{7-8}\cline{9-10}

\textbf{Answer model} & {\textbf{1-hop}}&{\textbf{2-hop}}&{\textbf{$\geq$3-hop}}&{\textbf{2-hop}}&{\textbf{$\geq$3-hop}}&{\textbf{2-hop}}&{\textbf{$\geq$3-hop}}&{\textbf{2-hop}}&{\textbf{$\geq$3-hop}}\\

\midrule
Qwen2.5-7b-instruct-1m  & \cellcolor{green!18}  84.10 & \cellcolor{red!20}65.25 & \cellcolor{red!30}60.25  & \cellcolor{green!21} 85.57& \cellcolor{green!32} 91.04&\cellcolor{green!31}91.09&\cellcolor{green!15}82.76&\cellcolor{red!14} 68.06 &\cellcolor{red!22}64.10\\

DeepSeek-R1-Distill-Qwen-32B& \cellcolor{green!30} 89.94 & \cellcolor{red!28}  61.03 &\cellcolor{red!18}  66.21 &\cellcolor{green!25} 87.51 & \cellcolor{green!29}89.47   &\cellcolor{green!26}88.23& \cellcolor{green!26}86.57 &\cellcolor{green!25}87.50& \cellcolor{green!28} 84.31 \\

Llama-3.3-70B-Instruct    & \cellcolor{green!29} 89.58 & \cellcolor{red!22}64.63 & \cellcolor{red!26}62.16 & \cellcolor{green!31}90.70&  \cellcolor{green!28}88.89 &\cellcolor{green!29}89.14& \cellcolor{green!28} 88.89 & \cellcolor{green!23} 86.81 & \cellcolor{green!19} 84.58 \\

GPT-3.5        & \cellcolor{green!31} 90.48 & \cellcolor{red!14} 68.18 &\cellcolor{red!12}  68.92&\cellcolor{green!29}88.37 & \cellcolor{green!14}86.11 & \cellcolor{green!33}91.35 & \cellcolor{green!13}82.28& \cellcolor{green!29}88.37 & \cellcolor{green!11}80.56\\

GPT-4o         & \cellcolor{green!32} 91.63 & \cellcolor{red!18}66.89 &\cellcolor{red!27} 62.16 & \cellcolor{green!31}90.58 &  \cellcolor{green!34}91.89 & \cellcolor{green!30}89.92&\cellcolor{green!28}88.89 &\cellcolor{green!28}89.28 & \cellcolor{green!12}81.08\\

\bottomrule
\end{tabular}
}
\caption{Evaluation of QA performance under different backbone configurations, varying both the Q-d and Answer models. “–” indicates no question decomposition applied.}
\vspace{-0.5em}
\label{tab:backbone-results}
\end{table*}

Table~\ref{tab:backbone-results} presents a detailed comparison across different backbone models under varying reasoning depths (1-hop, 2-hop, and 3-hop), both with and without question decomposition (Q-d). The results clearly demonstrate the critical role of question decomposition in enhancing multi-hop reasoning performance. Without decomposition, even strong models like GPT-3.5 and GPT-4o struggle on 3-hop questions, achieving only 68.92\% and 62.16\% accuracy respectively. However, when equipped with our question decomposition module (e.g., GPT-4o as the Q-d model), their performance improves substantially, reaching 86.11\% and 91.89\% respectively. This trend holds consistently across other backbone models and hop depths, confirming that decomposition provides a structured representation of complex questions—such as explicit constraints and a focused main query—that significantly reduces reasoning ambiguity.

More importantly, these results reveal that our framework reduces the dependency on the intrinsic reasoning capacity of the backbone LLM. For instance, the performance gap between smaller, instruction-tuned models (e.g., Qwen2.5-7B) and larger proprietary models (e.g., GPT-4o) is significantly narrowed or even eliminated when question decomposition is applied. This suggests that our framework acts as a form of “reasoning scaffolding,” allowing even lightweight models to perform competitively on complex, multi-table QA tasks. Furthermore, the decomposition-based pipeline enables better modularity and control in system design, as reasoning steps are made explicit and interpretable.

In sum, the consistent improvements observed across all reasoning depths and model configurations underscore the dual benefits of our approach: (1) improved reasoning accuracy through structured question understanding, and (2) reduced reliance on large-scale proprietary models, thereby enhancing the practicality and scalability of multi-table QA in real-world applications.

%% file: Appendix.tex
\section{Prompts}
\label{sec:appendix_prompt}
\subsection{Question decomposition}
We use the following prompt to ask LLMs to decompose a question into (conditions, main question).

{\fontfamily{cmtt}\selectfont 
\noindent You are a helpful assistant for question understanding and decomposition.
\noindent Your task is to rewrite a complex question into:
- A list of \textbf{conditions} (such as case ID, vehicle model, etc.)
- A \textbf{main question} that describes what needs to be answered

\noindent Return the result in the following format:

\noindent Condition 1: ...\\
\noindent Condition 2: ...\\
...\\
\noindent Main Question: ...\\

\noindent Example: \\
\noindent Question: For the vehicle model Bolt EV, were any active safety systems activated in case 20335? \\
Answer: \\
Condition 1: the vehicle model is Bolt EV\\
Condition 2: case ID is 20335\\
Main Question: were any active safety systems activated?\\

\noindent Example: \\
Question: In case 17390, what was the model year of the vehicle with most severe damage to the front? \\
Answer: \\
Condition 1: Case ID is 17390\\
Condition 2: The damage to the front is the most severe\\
Main Question: What was the model year of the vehicle?\\

\noindent Now decompose this question:
}
\subsection{Answer generation on CISS}
We use the following prompt to ask LLMs to generate answers for questions on CISS dataset:

{\fontfamily{cmtt}\selectfont

Question: What was the direction of travel of Vehicles in CASE 13803?

In case 13803, the crash summary is: Vehicle 1 and vehicle 2 were traveling east approaching an intersection. Vehicle 2 was stopped ahead of Vehicle 1. Vehicle 1's front plane contacted Vehicle 2 rear plane.
The number of in-transport vehicles is: 2

...

**Please read all the information provided above carefully and focus on question related information, answer the above question:**
}

\subsection{Attributes select for SQL}

We use the following prompt to ask LLMs to select attributes from tables:

{\fontfamily{cmtt}\selectfont You are a helpful assistant for aligning natural language question phrases with database variables.\\

\noindent You are given:\\
1. A database schema table with sheet$\_$name (table), column$\_$name (field), and key$\_$words (keywords associated with the field).\\
2. A natural language question.

\noindent Your task is:
Split the question into key semantic fragments (e.g., person name, gender, event, year, etc.).
For each fragment, return the most relevant column$\_$name and its corresponding sheet$\_$name based on the key$\_$words.
Ensure that each value in the output matches the database **exactly**, including full names, suffixes (e.g., "II", "Jr."), and punctuation (e.g., commas).

\noindent Tips:\\
Do not shorten names (e.g., avoid "Michael Phelps").\\
You must ...
...

\noindent Output format:\\

\noindent [\\
    \noindent "fragment": "...",\\
    \noindent "column$\_$name": "...", //...\\
    "gender", not "person.gender") //...\\
    "sheet$\_$name": "..."  //...\\
] \\
}

\subsection{SQL generation}
We use the following prompt to ask LLMs to generate SQLs:

{\fontfamily{cmtt}\selectfont  
\noindent Given the following reasoning paths and natural language question, generate the corresponding SQL query. 
 
\noindent Note: Some paths may be irrelevant to answering the question. Please carefully identify and ignore unrelated paths.

\noindent When assigning values to variables in the WHERE or SELECT clause, always prioritize using the variable at the **start of the path**, not internal IDs or primary keys. 

\noindent Only join tables that contribute directly to the SELECT, WHERE, or GROUP BY clauses. Avoid unnecessary joins (e.g., `games$\_$competitor`) if the question does not involve participation, age, or game-specific attributes. Unused joins may introduce incorrect row duplication.

...

\noindent Example:\\

\noindent Path 1: full$\_$name[person], id[person], person$\_$id[games$\_$competitor], id[games$\_$competitor]\\

\noindent Path 2: games$\_$name[games], id[games], games$\_$id[games$\_$competitor], id[games$\_$competitor]\\

\noindent Path 3: age[games$\_$competitor],  id[games$\_$competitor]  \\

\noindent Question:  
What was the name of the Olympic game that John Aalberg took part in when he was 31?\\

\noindent ```sql
SELECT games.games$\_$name
FROM person
JOIN games$\_$competitor ON person.id = games$\_$competitor.person$\_$id
JOIN games ON games.id = games$\_$competitor.games$\_$id
JOIN competitor$\_$event ON games$\_$competitor.id = competitor$\_$event.competitor$\_$id
WHERE person.full$\_$name = 'John Aalberg' AND games$\_$competitor.age = 31; 
```

}

\section{Case Study}
\subsection{1-hop question}
The following example highlights how our graph enhanced retrieval process. While the attribute CRASH.MANCOLL provides the target answer field, the graph also links to CRASH.SUMMARY, which offers contextual evidence describing the crash scenario. By incorporating this additional information, the model generates not only the correct answer but also a reliable explanation, making the response more trustworthy and grounded in the original data.

{\fontfamily{cmtt}\selectfont 

\noindent Question: What was the manner of the collision in CASE 6916?\\

\noindent Attributes retrieval result: MANCOLL[CRASH], CASEID[CRASH]\\

\noindent Graph searching result:\\
Path 1: MANCOLL[CRASH], SUMMARY[CRASH], CASEID[CRASH]\\

\noindent Information extraction result: In case 6916, the crash summary is: Vehicle \#1 was traveling west on an interstate highway, departed the road to the right, and impacted a guardrail with the front plane. \\
The MANCOLL is: Not Collision with Vehicle in Transport.\\

\noindent Answer: In case 6916, the manner of the collision is: Not Collision with Vehicle in Transport. This means that Vehicle \#1 did not collide with another vehicle in transport; instead, it departed the road and collided with a guardrail.}

\subsection{2-hop question}
This example demonstrates how the schema graph enhances the robustness of the multi-table QA pipeline. In this case, the attribute-level RAG module successfully retrieves \texttt{DVLONG} (longitudinal speed) but overlooks \texttt{DVTOTAL} (total speed), which is also essential to answer the question. However, our schema graph captures the semantic dependency between \texttt{DVLONG} and \texttt{DVTOTAL} by modeling them as connected nodes within the same table and linking them through shared keys such as \texttt{VEHNO} and \texttt{CASEID}. As a result, both attributes are extracted during graph traversal, effectively compensating for the RAG omission and ensuring that the model obtains complete information for accurate answering.

{\fontfamily{cmtt}\selectfont 

\noindent Question: For the case 27187, for the vehicle model Grand Prix, what is the speed of the vehicle?

\noindent Attributes retrieval result: DVLONG[GV], Model[VPICDECODE], CASEID[CRASH]\\

\noindent Graph searching result: \\
Path 1: DVTOTAL[GV], DVLONG[GV], VEHNO[GV], CASEID[GV]\\
Path 2: Model[VPICDECODE], VEHNO[VPICDECODE], CASEID[VPICDECODE]
Path 3: CASEID[VPICDECODE], CASEID[GV] \\

\noindent Information extraction result: In case 27187, for vehicle NO.2, the total speed is: \noindent Actual value 38, the longitudinal speed is: Actual value -37
\noindent In case 27187, for vehicle NO.2, the model is: Grand Prix\\
...\\

\noindent Answer: The total speed of the vehicle is 38.}
\subsection{3-hop question}
This 3-hop question requires reasoning over multiple conditions: identifying the correct occupant based on age (25 years), gender (female), and belt use (lap and shoulder belt), and then determining which body region made contact with the left door panel. While attribute-level retrieval alone may struggle to consistently extract all relevant fields, our schema graph enables the model to follow structured paths linking semantic and relational dependencies across tables. 

{\fontfamily{cmtt}\selectfont 
\noindent In case 16215, for the driver who is a 25-year-old female using a lap and shoulder belt, which body region contacted the left door panel?\\

\noindent Attributes retrieval result: ROLE[OCC], AGE[OCC], BELTUSE[OCC], BODYREGION[OCCONTACT], CONTAREA [OCCONTACT], CASEID[CRASH]\\

\noindent Graph searching result: \\
\noindent Path 1: BELTUSE[OCC], ROLE[OCC], OCCNO[OCC], VEHNO[OCC], CASEID[OCC]\\
\noindent Path 2: BODYREGION[OCCONTACT], CONTAREA[OCCONTACT], OCCNO[OCCONTACT], VEHNO[OCCONTACT], CASEID[OCCONTACT]\\
\noindent Path 3: AGE[OCC], ROLE[OCC], OCCNO[OCC], VEHNO[OCC], CASEID[OCC]
\noindent Path 3: CASEID[OCC], CASEID[OCCONTACT]
\\

\noindent Information extraction result: In case 16215, for vehicle NO.2, occupant NO.1, the occupant's role is: Driver, the belt-use determination is: Lap and shoulder belt.\\
In case 16215, for vehicle NO.2, occupant NO.1, the vehicle component the occupant is suspected to have contacted  is: Left rear upper quadrant, the CONTACTED COMPONENT AREA is: Left Door Panel, the the suspected body region that made contact to the Contacting Component is: Elbow - Left.\\
In case 16215, for vehicle NO.2, occupant NO.1, the occupant's role is: Driver, the occupant's age is: Actual age 25 years.\\

\noindent Answer: In case 16215, the left shoulder was the body region that contacted the left door panel.
}

\subsection{Fail cases}
These failure cases describe a key limitation of LLM-based reasoning in domain-specific contexts. Although the graph-based system successfully extracted all relevant information from tables, the LLM failed to aggregate these information appropriately due to a lack of domain-specific knowledge.

{\fontfamily{cmtt}\selectfont 
\noindent Question: What was the general area of damage on the first vehicle in CASE 18765?\\

\noindent Information extraction result: In case 18765, for event NO.1, the general damage area of first vehicle is: Front\\
In case 18765, for event NO.2, the general damage area of the first vehicle is: Back/Truck Back\\
In case 18765, for event NO.3, the general damage area of the first vehicle is: Front\\

\noindent Answer: The general damage area on the first vehicle (V1) in case 18765 was in the Front.\\
\noindent Ground true answer: The general area of damage on the first vehicle (V1) in CASE 18765 includes both the front and the back. \\}

This above example demonstrates a key limitation: the model failed to consolidate multiple related mentions (e.g., "Front", "Back/Truck Back") into a unified interpretation, defaulting to the first mention instead. \\

{\fontfamily{cmtt}\selectfont
\noindent Question: Was the vehicle equipped with any safety systems in CASE 31120?\\

\noindent Answer:In Case 31120, the vehicles (NO.1 and NO.2) were equipped with some safety systems, which are systems designed to prevent accidents or help reduce severity. From the information provided, the active safety systems available are:
 Automatic Crash Notification (Yes for both Vehicle NO.1 and Vehicle NO.2)\\
 
\noindent Information extraction result:In case 31120, for vehicle NO.1, the avoidance equipment to check is: Automatic Crash Notification, the availability condition is: Yes\\
In case 31120, for vehicle NO.1, the avoidance equipment to check is: Daytime Running Lamps, the availability condition is: Yes\\
In case 31120, for vehicle NO.1, the avoidance equipment to check is: Rearview Video System, the availability condition is: Yes\\
In case 31120, for vehicle NO.2, the avoidance equipment to check is: Automatic Crash Notification, the availability condition is: Yes
...

\noindent Ground true answer:  Yes, in case 31120, both Vehicle No.1 and Vehicle No.2 were equipped with safety systems. Specifically, they both had the following safety systems available:

\noindent - Automatic Crash Notification\\
\noindent - Daytime Running Lamps\\
\noindent - Rearview Video System\\

Additionally, Vehicle No. 2 also had Advanced Lighting available."\\
}

The answer is incorrect due to the language model's implicit assumption about the definition of safety systems. The model misclassified Daytime Running Lamps and Rearview Video System as not being part of the safety systems, likely due to its general knowledge. However, the table explicitly includes it under active safety features, leading to an incomplete answer.

%% file: emnlp2021.bbl
\begin{thebibliography}{31}
\expandafter\ifx\csname natexlab\endcsname\relax\def\natexlab#1{#1}\fi

\bibitem[{AI(2024)}]{meta2024llama3}
Meta AI. 2024.
\newblock \href {https://www.llama.com/models/llama-3/} {Llama 3: Open foundation and instruction-tuned models}.

\bibitem[{Cai et~al.(2022)Cai, Li, Hui, Yang, Li, Li, Cao, Li, Huang, Si et~al.}]{cai2022star}
Zefeng Cai, Xiangyu Li, Binyuan Hui, Min Yang, Bowen Li, Binhua Li, Zheng Cao, Weijie Li, Fei Huang, Luo Si, et~al. 2022.
\newblock Star: Sql guided pre-training for context-dependent text-to-sql parsing.
\newblock In \emph{Findings of the Association for Computational Linguistics: EMNLP 2022}, pages 1235--1247.

\bibitem[{Cao et~al.(2024)Cao, Zheng, Fan, Zhang, Chen, and Bai}]{cao2024rsl}
Zhenbiao Cao, Yuanlei Zheng, Zhihao Fan, Xiaojin Zhang, Wei Chen, and Xiang Bai. 2024.
\newblock Rsl-sql: Robust schema linking in text-to-sql generation.
\newblock \emph{arXiv preprint arXiv:2411.00073}.

\bibitem[{Chen et~al.(2024)Chen, Zhang, and Roth}]{chen-etal-2024-table}
Peter~Baile Chen, Yi~Zhang, and Dan Roth. 2024.
\newblock \href {https://arxiv.org/abs/2404.09889} {Is table retrieval a solved problem? exploring join-aware multi-table retrieval}.
\newblock \emph{arXiv preprint arXiv:2404.09889}.

\bibitem[{DeepSeek-AI(2025)}]{deepseekai2025deepseekr1incentivizingreasoningcapability}
DeepSeek-AI. 2025.
\newblock \href {http://arxiv.org/abs/2501.12948} {Deepseek-r1: Incentivizing reasoning capability in llms via reinforcement learning}.

\bibitem[{Herzig et~al.(2020)Herzig, Nowak, Mueller, Piccinno, and Eisenschlos}]{herzig2020tapas}
Jonathan Herzig, Pawel~Krzysztof Nowak, Thomas Mueller, Francesco Piccinno, and Julian Eisenschlos. 2020.
\newblock Tapas: Weakly supervised table parsing via pre-training.
\newblock In \emph{Proceedings of the 58th Annual Meeting of the Association for Computational Linguistics}, pages 4320--4333.

\bibitem[{Jin et~al.(2022)Jin, Siebert, Li, and Chen}]{jin2022survey}
Nengzheng Jin, Joanna Siebert, Dongfang Li, and Qingcai Chen. 2022.
\newblock A survey on table question answering: recent advances.
\newblock In \emph{China Conference on Knowledge Graph and Semantic Computing}, pages 174--186. Springer.

\bibitem[{Li et~al.(2024)Li, Hui, Qu, Yang, Li, Li, Wang, Qin, Geng, Huo et~al.}]{li2024can}
Jinyang Li, Binyuan Hui, Ge~Qu, Jiaxi Yang, Binhua Li, Bowen Li, Bailin Wang, Bowen Qin, Ruiying Geng, Nan Huo, et~al. 2024.
\newblock Can llm already serve as a database interface? a big bench for large-scale database grounded text-to-sqls.
\newblock \emph{Advances in Neural Information Processing Systems}, 36.

\bibitem[{Liu et~al.()Liu, Chen, Guo, Ziyadi, Lin, Chen, and Lou}]{liutapex}
Qian Liu, Bei Chen, Jiaqi Guo, Morteza Ziyadi, Zeqi Lin, Weizhu Chen, and Jian-Guang Lou.
\newblock Tapex: Table pre-training via learning a neural sql executor.
\newblock In \emph{International Conference on Learning Representations}.

\bibitem[{Ma et~al.(2022)Ma, Cheng, Liu, Nyberg, and Gao}]{ma2022open}
Kaixin Ma, Hao Cheng, Xiaodong Liu, Eric Nyberg, and Jianfeng Gao. 2022.
\newblock Open domain question answering with a unified knowledge interface.
\newblock In \emph{Proceedings of the 60th Annual Meeting of the Association for Computational Linguistics (Volume 1: Long Papers)}, pages 1605--1620.

\bibitem[{Maamari et~al.(2024)Maamari, Abubaker, Jaroslawicz, and Mhedhbi}]{maamari2024death}
Karime Maamari, Fadhil Abubaker, Daniel Jaroslawicz, and Amine Mhedhbi. 2024.
\newblock The death of schema linking? text-to-sql in the age of well-reasoned language models.
\newblock \emph{arXiv preprint arXiv:2408.07702}.

\bibitem[{Nan et~al.(2022)Nan, Hsieh, Mao, Lin, Verma, Zhang, Kry{\'s}ci{\'n}ski, Schoelkopf, Kong, Tang et~al.}]{nan2022fetaqa}
Linyong Nan, Chiachun Hsieh, Ziming Mao, Xi~Victoria Lin, Neha Verma, Rui Zhang, Wojciech Kry{\'s}ci{\'n}ski, Hailey Schoelkopf, Riley Kong, Xiangru Tang, et~al. 2022.
\newblock Fetaqa: Free-form table question answering.
\newblock \emph{Transactions of the Association for Computational Linguistics}, 10:35--49.

\bibitem[{{{National Highway Traffic Safety Administration}}(2024)}]{nhtsa-2024-ciss}
{{National Highway Traffic Safety Administration}}. 2024.
\newblock \href {https://www.nhtsa.gov/file-downloads?p=nhtsa/downloads/CISS/} {Crash investigation sampling system (ciss)}.
\newblock Accessed: 2025-05-07.

\bibitem[{OpenAI(2023)}]{openai2023gpt35}
OpenAI. 2023.
\newblock \href {https://platform.openai.com/docs/models/gpt-3-5} {Gpt-3.5 series}.

\bibitem[{OpenAI(2024{\natexlab{a}})}]{openai2024gpt4o}
OpenAI. 2024{\natexlab{a}}.
\newblock \href {https://openai.com/index/gpt-4o} {Gpt-4o technical report}.

\bibitem[{OpenAI(2024{\natexlab{b}})}]{openai2024embedding}
OpenAI. 2024{\natexlab{b}}.
\newblock \href {https://platform.openai.com/docs/guides/embeddings} {text-embedding-3-small model}.

\bibitem[{Pal et~al.(2022)Pal, Kanoulas, and Rijke}]{pal2022parameter}
Vaishali Pal, Evangelos Kanoulas, and Maarten Rijke. 2022.
\newblock Parameter-efficient abstractive question answering over tables or text.
\newblock In \emph{Proceedings of the Second DialDoc Workshop on Document-grounded Dialogue and Conversational Question Answering}, pages 41--53.

\bibitem[{Pal et~al.(2023)Pal, Yates, Kanoulas, and de~Rijke}]{pal2023multitabqa}
Vaishali Pal, Andrew Yates, Evangelos Kanoulas, and Maarten de~Rijke. 2023.
\newblock Multitabqa: Generating tabular answers for multi-table question answering.
\newblock \emph{arXiv preprint arXiv:2305.12820}.

\bibitem[{Singhal et~al.(2025)Singhal, Tu, Gottweis, Sayres, Wulczyn, Amin, Hou, Clark, Pfohl, Cole-Lewis et~al.}]{singhal2025toward}
Karan Singhal, Tao Tu, Juraj Gottweis, Rory Sayres, Ellery Wulczyn, Mohamed Amin, Le~Hou, Kevin Clark, Stephen~R Pfohl, Heather Cole-Lewis, et~al. 2025.
\newblock Toward expert-level medical question answering with large language models.
\newblock \emph{Nature Medicine}, pages 1--8.

\bibitem[{Srivastava et~al.(2024)Srivastava, Malik, Gupta, Ganu, and Roth}]{srivastava2024evaluating}
Pragya Srivastava, Manuj Malik, Vivek Gupta, Tanuja Ganu, and Dan Roth. 2024.
\newblock Evaluating llms' mathematical reasoning in financial document question answering.
\newblock \emph{arXiv preprint arXiv:2402.11194}.

\bibitem[{Wang et~al.(2024)Wang, Wang, Zhang, Zhao, Yang, and Guo}]{wang2024multi}
Bing Wang, Chunhao Wang, Xingpeng Zhang, Chunlan Zhao, Xiaoling Yang, and Kuan Guo. 2024.
\newblock Multi-table question answering method based on correlation evaluation and precomputed cube.
\newblock In \emph{International Conference on Knowledge Science, Engineering and Management}, pages 393--405. Springer.

\bibitem[{Wang and Yu(2025)}]{wang-yu-2025-iquest}
Shuai Wang and Yinan Yu. 2025.
\newblock \href {https://doi.org/10.18653/v1/2025.acl-long.760} {i{QUEST}: An iterative question-guided framework for knowledge base question answering}.
\newblock In \emph{Proceedings of the 63rd Annual Meeting of the Association for Computational Linguistics (ACL)}, pages 15616--15628.

\bibitem[{Wu et~al.(2025{\natexlab{a}})Wu, Yang, Li, Ji, Okumura, and Zhang}]{wu-etal-2025-mmqa}
Jian Wu, Linyi Yang, Dongyuan Li, Yuliang Ji, Manabu Okumura, and Yue Zhang. 2025{\natexlab{a}}.
\newblock {MMQA}: Evaluating {LLMs} with multi-table multi-hop complex questions.
\newblock In \emph{Proceedings of the Thirteenth International Conference on Learning Representations (ICLR)}.

\bibitem[{Wu et~al.(2025{\natexlab{b}})Wu, Yang, Chai, Zhang, Liu, Du, Liang, Shu, Cheng, Sun et~al.}]{wu2025tablebench}
Xianjie Wu, Jian Yang, Linzheng Chai, Ge~Zhang, Jiaheng Liu, Xeron Du, Di~Liang, Daixin Shu, Xianfu Cheng, Tianzhen Sun, et~al. 2025{\natexlab{b}}.
\newblock Tablebench: A comprehensive and complex benchmark for table question answering.
\newblock In \emph{Proceedings of the AAAI Conference on Artificial Intelligence}, volume~39, pages 25497--25506.

\bibitem[{Xie et~al.(2025)Xie, Xu, Zhao, and Guo}]{xie2025opensearch}
Xiangjin Xie, Guangwei Xu, Lingyan Zhao, and Ruijie Guo. 2025.
\newblock Opensearch-sql: Enhancing text-to-sql with dynamic few-shot and consistency alignment.
\newblock \emph{arXiv preprint arXiv:2502.14913}.

\bibitem[{Yang et~al.(2025)Yang, Yu, Li, Liu, Huang, Huang, Jiang, Tu, Zhang, Zhou, Lin, Dang, Yang, Yu, Li, Sun, Zhu, Men, He, Xu, Yin, Yu, Qiu, Ren, Yang, Li, Xu, and Zhang}]{qwen2.5}
An~Yang, Bowen Yu, Chengyuan Li, Dayiheng Liu, Fei Huang, Haoyan Huang, Jiandong Jiang, Jianhong Tu, Jianwei Zhang, Jingren Zhou, Junyang Lin, Kai Dang, Kexin Yang, Le~Yu, Mei Li, Minmin Sun, Qin Zhu, Rui Men, Tao He, Weijia Xu, Wenbiao Yin, Wenyuan Yu, Xiafei Qiu, Xingzhang Ren, Xinlong Yang, Yong Li, Zhiying Xu, and Zipeng Zhang. 2025.
\newblock Qwen2.5-1m technical report.
\newblock \emph{arXiv preprint arXiv:2501.15383}.

\bibitem[{Ye et~al.(2023)Ye, Hui, Yang, Li, Huang, and Li}]{ye2023large}
Yunhu Ye, Binyuan Hui, Min Yang, Binhua Li, Fei Huang, and Yongbin Li. 2023.
\newblock Large language models are versatile decomposers: Decomposing evidence and questions for table-based reasoning.
\newblock In \emph{Proceedings of the 46th international ACM SIGIR conference on research and development in information retrieval}, pages 174--184.

\bibitem[{Zhang et~al.(2024{\natexlab{a}})Zhang, Yue, Li, and Sun}]{zhang2024tablellama}
Tianshu Zhang, Xiang Yue, Yifei Li, and Huan Sun. 2024{\natexlab{a}}.
\newblock Tablellama: Towards open large generalist models for tables.
\newblock In \emph{Proceedings of the 2024 Conference of the North American Chapter of the Association for Computational Linguistics: Human Language Technologies (Volume 1: Long Papers)}, pages 6024--6044.

\bibitem[{Zhang et~al.(2024{\natexlab{b}})Zhang, Luo, Zhang, Ma, Zhang, Li, Li, Yao, Xu, Zhou et~al.}]{zhang2024tablellm}
Xiaokang Zhang, Sijia Luo, Bohan Zhang, Zeyao Ma, Jing Zhang, Yang Li, Guanlin Li, Zijun Yao, Kangli Xu, Jinchang Zhou, et~al. 2024{\natexlab{b}}.
\newblock Tablellm: Enabling tabular data manipulation by llms in real office usage scenarios.
\newblock \emph{arXiv preprint arXiv:2403.19318}.

\bibitem[{Zhang et~al.(2024{\natexlab{c}})Zhang, Sun, Wang, Nie, Ma, Sun, and Li}]{zhang2024large}
Zijian Zhang, Yujie Sun, Zepu Wang, Yuqi Nie, Xiaobo Ma, Peng Sun, and Ruolin Li. 2024{\natexlab{c}}.
\newblock Large language models for mobility in transportation systems: A survey on forecasting tasks.
\newblock \emph{arXiv preprint arXiv:2405.02357}.

\bibitem[{Zhu et~al.(2021)Zhu, Lei, Huang, Wang, Zhang, Lv, Feng, and Chua}]{zhu2021tat}
Fengbin Zhu, Wenqiang Lei, Youcheng Huang, Chao Wang, Shuo Zhang, Jiancheng Lv, Fuli Feng, and Tat-Seng Chua. 2021.
\newblock Tat-qa: A question answering benchmark on a hybrid of tabular and textual content in finance.
\newblock In \emph{Proceedings of the 59th Annual Meeting of the Association for Computational Linguistics and the 11th International Joint Conference on Natural Language Processing (Volume 1: Long Papers)}. Association for Computational Linguistics.

\end{thebibliography}
